\documentclass[letterpaper, 10 pt, conference]{ieeeconf}  

\IEEEoverridecommandlockouts                              

\usepackage{graphics} 
\usepackage{epsfig} 
\usepackage{mathptmx} 
\usepackage{times} 
\usepackage{amsmath} 
\usepackage{amssymb}  
\usepackage{multirow}
\usepackage{color}
\usepackage{hhline}
\usepackage{color,soul}
\usepackage{adjustbox}
\usepackage{tabularx}
\usepackage{float}

\title{\LARGE \bf
Focused Blind Switching Manipulation Based on Constrained and Regional Touch States of Multi-Fingered Hand Using Deep Learning
}

\author{S. Funabashi$^{1}$, A. Hiramoto$^{1}$, N. Chiba$^{2}$, A. Schmitz$^{1}$, S. Kulkarni$^{1}$ and T. Ogata$^{1}$ 
\thanks{This research was supported by Grant-in-Aid for Early-Career Scientists (KAKENHI) from Japan Society for the Promotion of Science (JSPS) with a grant number of 1167393 and Moonshot R\&D with a grant number of JPMJMS2031.}
\thanks{$^{1}$ The authors are with Waseda University, Okubo 3-4-1, Shinjuku, Tokyo 169-8555, Japan.
$^{2}$ The author is with Osaka University, Cybermedia Center, Machikaneyama 1-32, Toyonaka, Osaka, Japan.
}
}

\begin{document}

\maketitle
\thispagestyle{empty}
\pagestyle{empty}

\begin{abstract}

To achieve a desired grasping posture (including object position and orientation), multi-finger motions need to be conducted according to the the current touch state. 
Specifically, when subtle changes happen during correcting the object state, not only proprioception but also tactile information from the entire hand can be beneficial. 
However, switching motions with high-DOFs of multiple fingers and abundant tactile information is still challenging. In this study, we propose a loss function with constraints of touch states and an attention mechanism for focusing on important modalities depending on the touch states. The policy model is AE-LSTM which consists of Autoencoder (AE) which compresses abundant tactile information and Long Short-Term Memory (LSTM) which switches the motion depending on the touch states. Motion for cap-opening was chosen as a target task which consists of subtasks of sliding an object and opening its cap. 
As a result, the proposed method achieved the best success rates with a variety of objects for real time cap-opening manipulation. Furthermore, we could confirm that the proposed model acquired the features of each subtask and attention on specific modalities.

\end{abstract}

\section{INTRODUCTION}
Humans can do difficult manipulation such as 
grasping an object and regrasping or reorienting it to achieve the final grasping posture
\cite{trend}. In this case, tactile feedback can be crucial. For example, humans roughly grab a key in a pocket without looking at it, and regrasp it to insert it to keyhole of a door. Furthermore, humans do regrasping again if the grasping position or orientation of the key is wrong to fit in the keyhole via tactile feedback. This is also observed in a variety of scenarios including using tools and manipulation of daily objects such as opening a cap. For robotic multi-fingered manipulation, it is still challenging especially to recognize touch or object states and plan different motions. 

    \begin{figure}[t]
      \centering
      \includegraphics[scale=0.26]{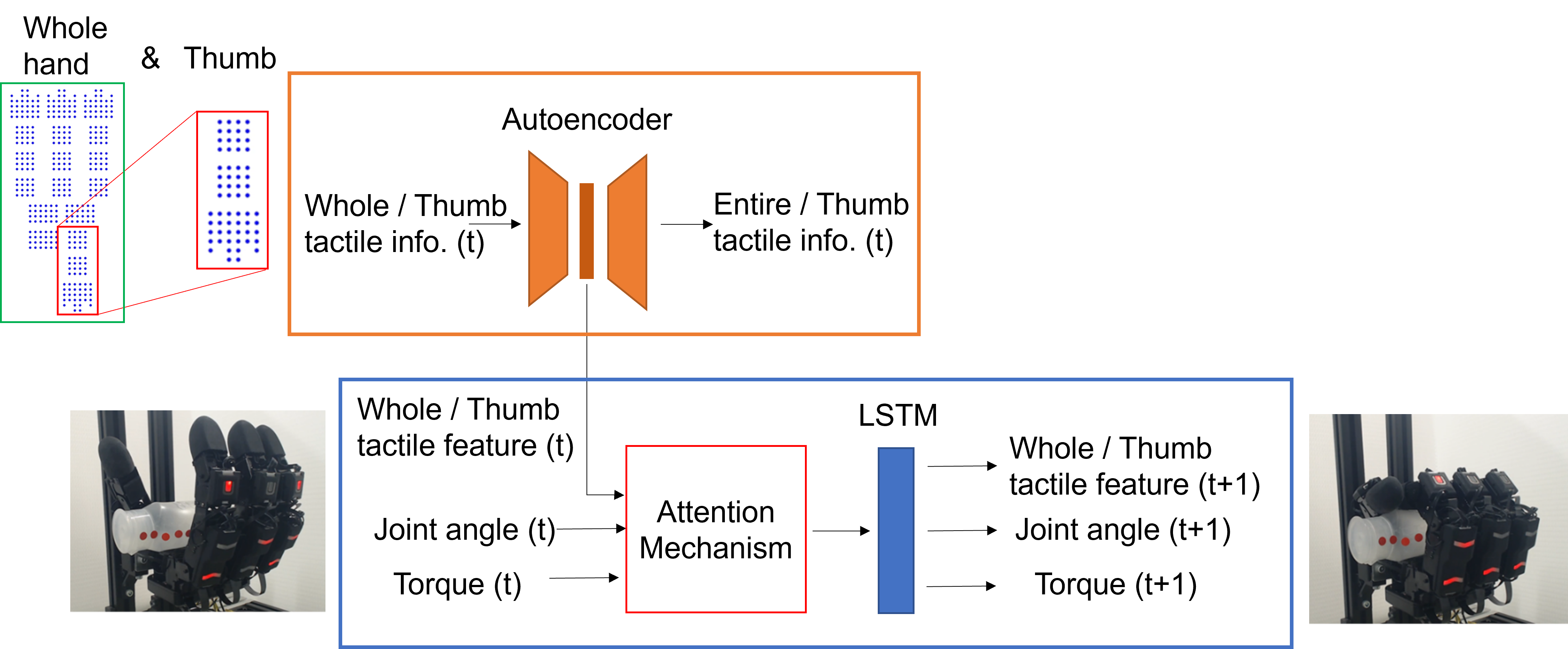}
      \caption{Schematic of the proposed motion-generating method. The model consists of AE and LSTM blocks to have an attention mechanism and handle time-series information.
      }
      \label{schematic}
   \end{figure}

In our previous work, temporal tactile information and parametric biases were utilized to switch to the desired motion with tactile feedback 
using deep imitation learning \cite{funabashirnn}. For achieving more dexterous tasks using a multi-fingered hand entirely, abundant tactile information was used to extract its features for generating motions \cite{funabashigcn}. Although it could achieve dexterous manipulation with extracted tactile features, it 
could not be adaptive to dexterous temporal manipulation such as switching to a different motion. 

\cite{putinbox} achieved the motions with recurrent neural networks 
with imposed constraints on the loss function to force the models to learn switching timing for achieving sequential sub-tasks. This method can be effective for the case of multi-fingered tactile tasks. 
However, it is still difficult to achieve multi-fingered sub-tasks because the sequential sub-tasks accumulate position and/or orientation errors of the grasped object, which eventually results in a failure of the whole manipulation. One of solutions when using deep learning is an attention mechanism
to focus on a specific input depending on the sub-task \cite{cooking}.

For these reasons, this study employs AE-LSTM in which an autoencoder (AE) extracts features from abundant tactile information and long short-term memory (LSTM) which handles temporal information of proprioception (joint and torque information) and tactile features. Furthermore, we implemented the constraints on a loss function mean squared error for training models of switching motions. We also implemented attention layer before LSTM layer to weigh modalities of input for the LSTM.
 Fig. \ref{schematic} shows a schematic of the proposed method.

Therefore, this paper presents these contributions: 
\begin{itemize}
\item Constraints on loss function to achieve multi-fingered switching motions for sub-tasks.
\item Attention mechanism to focus on important modalities for each sub-task to achieve fine manipulation. 
\item Achieved cap-opening manipulation with untrained objects and initial object positions.
\item Analysis of how the proposed method acquired the features of sub-tasks by PCA and the corresponding 
modalities by attention values which can be relevant to achievements of successful switching motions.
\end{itemize}

\section{Related Work}
\subsection{Switching Motion for Multi-Fingered Manipulation}
To reach a final grasping posture, 
some intermediate postures/motions to switch can be necessary due to kinematic constraint of robot hand shape or force constraint for keeping a stable grasp with an object \cite{twendyregrasp}. 
Some studies developed a robotic manipulator which can change a grasping posture with its mechanism depending on manipulation  scenarios \cite{rigidcontinuum}\cite{heavy}. \cite{contactmode} achieved motion planning of contact-rich manipulation based on contact mode. 
Although it can switch grasping postures/motions, how to recognize when to change them is important. \cite{policyblend} achieved to blend policies with modalities including tactile information.
In this case, one motion is blended with another specific motion and thus it can be difficult to switch motions for different touch state. Reinforcement learning is 
a way to achieve dexterous manipulation and change finger motions depending on grasping states \cite{living}\cite{rotate3d}. However 
most of them require tremendous training data \cite{openai}. 
Some studies focused on sim2real to approach dexterous manipulation \cite{eureka} and using tactile information \cite{toru}. On the other hand, \cite{fingergaiting} achieved motion generation with a multi-fingered hand by visual feedback without a cost-taking deep learning method. Importantly, those dexterous manipulation studies actually did not achieve fine manipulation which can require capturing subtle change of touch states to plan a next motion. Therefore, how to process such states for the manipulation should be investigated. \cite{putinbox} achieved switching motions in-between sub-tasks by imposing constraints on the loss function by focusing a change in visual sensing information. 
This method can be also useful for recognition of subtle touch states on an entire multi-fingered hand through tactile information. 
 
\subsection{Attention Mechanism for Multi-Fingered Sub-Tasks}
Even if switching motions is possible, it is also important perspective that beneficial modalities or sensory information can be changed depending on the sub-tasks. 
There are many ways to focus on modalities; e.g. spatial key coordinates \cite{vit} or temporal information \cite{bert}. Transformer is popular 
with its attention mechanism these days \cite{transformer}. Many studies have addressed robotic manipulation tasks \cite{gnnmanip}. \cite{vistactransformer} used vision and tactile information and predicted grasp outcomes. In \cite{objpack}, grasping areas on target objects were specified and achieved stable grasping of cluttered objects \cite{safegrasp}. One study showed which robot arm should be focused on and achieved dexterous manipulation tasks. Tactile sensing tasks were also achieved \cite{texture}\cite{sct-cnn}. In spite of powerful processing of transformer-based neural network models, they usually require a huge sized dataset. This is problematic especially for tactile manipulation because tactile senors easily get worn and break as they directly touch manipulated objects. 
Moreover, 
focusing on modalities including proprioception and tactile information depending on the sub-tasks can be more important as differences in each sub-task are subtle but crucial to achieve fine multi-fingered manipulation. Therefore, the attention mechanism for modalities proposed by \cite{cooking} is suitable for our target tasks.

\section{Proposed Method}
   \begin{figure*}[t]
      \centering
      \includegraphics[scale=0.48]{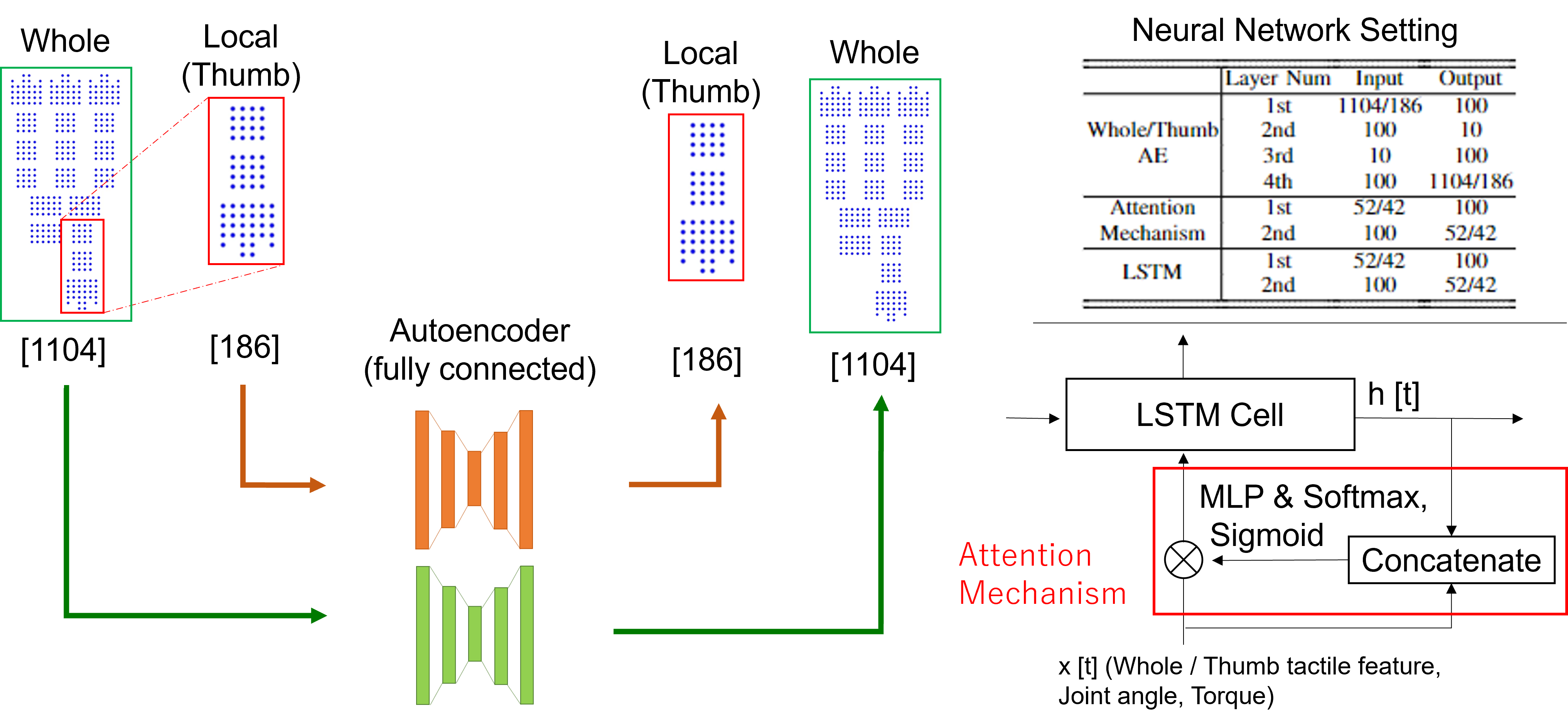}
      \caption{Detailed proposed AE-LSTM architecture. The AE is prepared for tactile information from the whole hand and local part of the hand (thumb in this study), respectively. The tactile features with joint and torque information are input to the attention layer so that the proposed network can focus on arbitrary modalities depending on the sub-tasks. The attention layer outputs the weighed features by the attention and they are input to the LSTM. The LSTM predicts the next time step of the sensor information (joint, torque and tactile features). The predicted joint information is used for controlling the Allegro Hand.}
      \label{proposedmodel}
   \end{figure*}


\subsection{Auto Encoder and Long-Short Term Memory}
As shown in Fig. \ref{schematic}, the proposed model combines an AE and an LSTM. The AE compresses high-dimensional tactile data into low-dimensional features of a size similar to that of the joint angle values and torque values. The LSTM generates motion by predicting future information through deep predictive learning. Such a two-stage model combining feature extractor and motion generation models is widely used in robot manipulation and can exhibit high generalization performance with reasonable training data \cite{cooking}.

As shown in Fig. \ref{proposedmodel}, we provided two AEs: one for obtaining tactile features of the entire finger and palm and another for obtaining local tactile features of only the thumb. This structure, also observed in prior research \cite{cooking}, aims to improve the success rate of tasks by separately providing tactile features of only the thumb, considering the role of the thumb in the task of cap-opening described in Section IV-A-2. 

We adopted the deep predictive learning approach as one of imitation learning methods, for motion generation model based on LSTM. The LSTM generates a next time step of data which includes not only joint information to control a robot hand but also sensory data such as tactile data. This enables motion generation models to reduce prediction error against actual environment resulting in achieving stable successful manipulation \cite{dpl}.

\subsection{Loop features on the latent space through constraints on the loss function}
During the training of the LSTM, 
the model should learn to focus on input data at the timing of switching motions and generate the switching motions.
The loss function used for training the model is the mean squared error, which aims to minimize the difference between the teaching data and the model's output. In addition to this, the proposed method introduces imposing constraints using the squared error of the hidden features of LSTM at the start and end of specific sub-tasks.

\begin{equation}
Loss=\sum_{t=0}^T\parallel  \hat(y)(t)-y(t)  \parallel ^2 + \gamma(h(T)-h(0))^2
\end{equation}

 where $T$ is the total number of time steps in the task, $\hat(y)(t)$ and $(y)(t)$ are the predicted values of the model and the teacher data respectively, $h(t)$ is the latent variable of the hidden layer in LSTM, and $\gamma$ is a parameter determining the strength of the constraint. This constraint on the loss function enables LSTM to learn to generate loop trajectories of motions on the latent space.
 By generating loop trajectories, the importance of input values at the timing of motion switching increases to emphasize the timings. It allows the model to learn to decide the next motion based on the state of the object or hand at that timing. Additionally, by fixing the joint angles at the time of motion switching, attention can be directed to other modalities such as tactile, which can capture touch or object states and are important for selecting motions.

\subsection{Attention Mechanism}
Within a sequence of motions to perform a task, the modalities of greater significance for each motion constantly change. Therefore, an attention mechanism is introduced to weigh each modality based on the past and present states of the motions. In this study, an attention mechanism is applied to the input values to LSTM with the purpose of focusing on modalities, as shown in Fig. \ref{proposedmodel}. The attention mechanism $A(t)$ for each modality uses the calculated value from the following equation:

\begin{equation}
A(t)=softmax(mlp(concatenate(h(t),x_in(t))))
\end{equation}

where $h(t)$ is the latent variable of the hidden layer in LSTM, $x_in(t)$ is the input data to the model, such as tactile features, joint angles, and torque, and mlp denotes a multilayer perceptron. The latent variables of LSTM and the input data are concatenated into a single column and transformed into attention for each modality by fully connected layers and the softmax function. The output dimension of the fully connected layer must match the size of input modalities. In this study, four modalities were used: whole and local tactile features, joint angles, and torque, hence the size of output layer is 4.

The input values $x_{lstm}(t)$ to LSTM use the weighted value obtained from the attention mechanism multiplied by the original input data, as shown in the following equation:

\begin{equation}
x_{lstm} = A(t) x_{in}(t)
\end{equation}

\section{Experiment Setting}

\subsection{Collecting Data}
\subsubsection{Hardware Setting}
An Allegro Hand, a multi-finger robot hand with four fingers was used. With 4 fingers with 4 DOF, 16 joints information was obtained. The uSkin sensor, a distributed tri-axial tactile sensor
was used. With 368 (sensor chips) * 3 axes, a total of 1104 measurements were used.

\subsubsection{Target Motion}

   
The experimental task chosen was the "cap-opening motion". The task involves multiple sub-tasks such as grasping the bottle, sliding it to the openable position, and opening the cap. To successfully complete the task, switching between these sub-tasks appropriately depending on the situation is necessary. Tactile feedback is crucial for determining whether the cap is in the openable position and confirming whether the cap has been opened after the opening motion, serving as important cues for switching between sub-tasks. Therefore, the cap-opening motion was chosen as the task to achieve appropriate switching of motions based on tactile feedback.

Furthermore, the cap-opening motion involves 
performing the motion at various initial positions, requiring generalization to left-right positions, shapes and sizes of objects. Additionally, recognizing the untrained size and position of the cap to generate the cap opening trajectory introduces various errors during motion generation that does not present in trained motions. Therefore, 
a deep predictive learning model that generates motions to reduce prediction errors depending on the situation is suitable for this task.

The motions were programmed by specifying the target joint angles and the time interval assigned to reach the posture of sub-tasks. The motion flow include the following motion patterns. 1. It performs grasping motion 
at first. 2. Then, it attempts cap-opening. If it is successful, the robot hand stops the motion. 3. If the cap does not open, the hand retracts the thumb and performs grasping motion again. 4. If the object is positioned to the left, the hand slides the object to the right and vice-versa. 5. After the sliding motion, it returns to a cap-opening motion. These sub-tasks were executed sequentially for data collection as training data. The data was cut into sub-tasks to train a model to learn each motion.

\begin{figure}[t]
      \centering
      \includegraphics[scale=0.27]{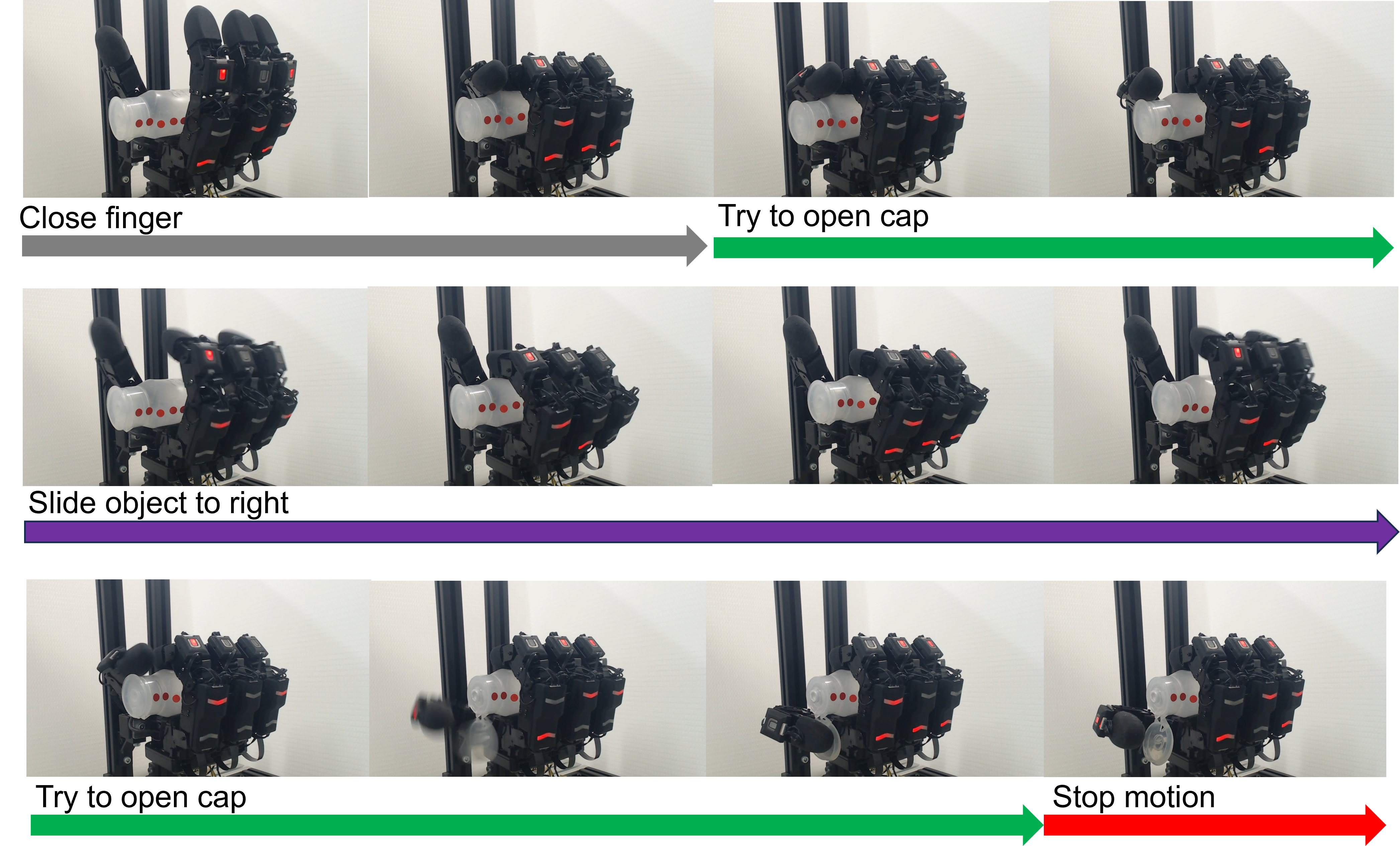}
      \caption{Target manipulation motion. Firstly, fingers are closed. Then the robot tries to open a cap. If the cap is opened, the robot stops, otherwise the robot slides the grasped object either right or left. Repeatedly, the robot tries to open the cap.}
      \label{manipulation}
   \end{figure}

To utilize the proposed method of imposing constraints on the loss function, the start and end values for modalities other than tactile features need to be similar for each task. Therefore, it is advantageous to align the joint angles at the start of the next sub-task and end of the current sub-task within a certain interval during the training motions based on the aforementioned grasping motions and motion flow. An example of actual training motions obtained through this motion design is shown in Fig. \ref{manipulation}

\subsubsection{Target Objects}
\begin{figure}[t]
      \centering
      \includegraphics[scale=0.27]{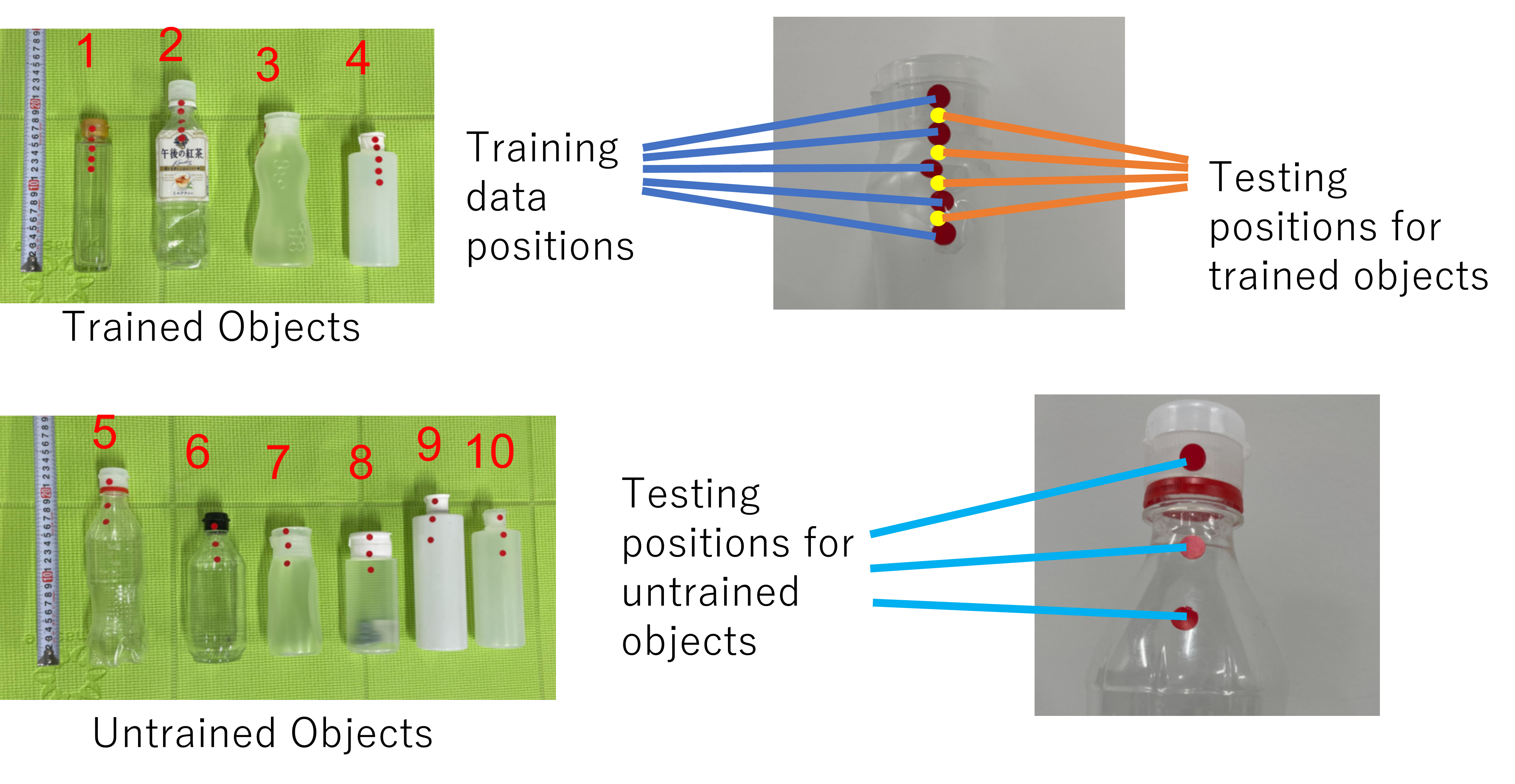}
      \caption{Target objects and initial grasping positions. Trained objects and positions: 5 daily objects were prepared. We put 5 red markers around a cap as an initial position for trained objects. The initial position is fixed to be the same position as the index finger. Testing positions were also prepared for an evaluation of real-time manipulation. The markers are 4 in-between the markers for training. Untrained objects: Ten objects were prepared for evaluating the generalization ability of the proposed method. Three red markers were put as initial positions for testing.}
      \label{datasetting}
   \end{figure}

The trained objects
and the untrained objects
are shown in Fig. \ref{datasetting}. To appropriately verify the generalization of the proposed method, various sizes, shapes, and materials of bottles with caps were prepared. For the trained objects, five markers were attached as initial positions
for collecting training data. The positions of the red markers were evenly spaced from the limit position 
where the Allegro Hand did not drop the object, to the end of the cap.

Since training data were collected for four types of bottles at five initial positions, with two trial ran at each position, the total number of training data is 40. As shown in Fig. \ref{datasetting}, the initial positions were fixed so that the center of the red markers overlaps with the index finger frame of the Allegro Hand. 


The training data was recorded at a sampling rate of 100 Hz. It was then re-sampled to 10~Hz and used as input to the models, since the tactile data from whole tactile sensors was fully updated at around 10~Hz. Additionally, the models were trained to predict the values 2 timesteps ahead (i.e., 0.2 seconds ahead). The input data were normalized to the range of 0.1 to 0.9. For joint angles and torque, scaling was performed independently for each joint (e.g., using the maximum and minimum values of joint j0 for scaling). For tactile information, scaling was performed independently for each patch and each axis (e.g., using the maximum and minimum values of the x-axis tactile information of Patch0 for scaling). Moreover, prior to scaling tactile information,
filtering of maximum and minimum values was performed. Tactile information obtained from uSkin can recognize subtle displacements, such as light touches or strokes. However, since these tactile information values are very small compared to displacements when tightly grasped, there is a problem that they disappear if they are simply normalized as raw data. Therefore, by performing filtering with the maximum value to 1000 and the minimum value to -1000 before scaling, loss of subtle tactile information was minimized.

To improve the generalization of the model, Gaussian noise following a normal distribution was added to the 
training data. The hyperparameters determining the strength of the noise were set for each input modality independently.

\subsection{Neural Network Settings}


Fig. \ref{proposedmodel} shows the model structure of the AE, the attention mechanism and the LSTM. 
All AEs in this study consisted of fully connected layers. The uSkin tactile sensor embedded in the Allegro Hand is a triaxial tactile sensor, capable of obtaining 1104 data points from 368 sensors. By flattening these tactile data into a single row and inputting them into the AE, compressed tactile latent features were obtained. Both the whole tactile information and the tactile information of the thumb were compressed to latent features of size 10. The input feature size to LSTM is 52 in total. Out of the 52 features, 10 features are of thumb tactile modality, 10 are of whole tactile modality, and 16 features each of joint angles and torque modalities. The attention mechanism employs two fully connected layers. The 4-dimensional weights outputted by the Softmax function are multiplied with each modality, including joint angles, torque, overall tactile features, and thumb tactile features, to provide weighted inputs to the LSTM.

In contrast, the baseline model without the attention mechanism utilized only thumb tactile features as input, resulting in 42 dimensions. As the 
whole tactile features on the entire hand also include the thumb's tactile information, the local tactile features of thumb were added to the input data only when the attention mechanism was used. Note that the attention can weigh each modality but it does not do so in one feature, i.e. the thumb tactile information in the whole tactile features.

The parameter values used for training all of the models was the same except of epochs. The models were trained for a maximum of 50,000 epochs, the models were trained for the epoch with the minimum loss on the validation data. The loss function was the mean squared error with the learning rate of 0.001. 

The weight values were assigned to loss corresponding to each input modality
Strong weight is applied to the finger base joint for sliding movements and to the thumb joint angles crucial for cap-opening movements. For cap-opening movements, there are two points of transition:

The proposed model was trained to switch motions appropriately by imposing restrictions on loss function for two timings listed below. 
\begin{itemize}
\item Timing of the finger motions after sliding or retracting the thumb.
\item Timing after attempting the cap-opening motion.
\end{itemize}
During the training, the loss function minimizes the loss on the timings. At the former timing, the options include sliding motions in either direction or three patterns of trained cap-opening motions. At the latter timing, the options include retracting the thumb motion or stopping the motion. Recognizing the state of the object or hand at these timings is essential for appropriate selection of finger motions for successful manipulation of cap-opening.

\section{Evaluation}
\subsection{Ablation Study with Proposed Model}

\setlength{\textfloatsep}{1pt}
\begin{table*}[t]
\caption{Success Rate of In-Hand Manipulation with Each Model}
\label{tab:3}
\centering
\tabcolsep = 3 pt
\renewcommand{\arraystretch}{1.0}
\begin{tabular}{l|cc|cc}
\hhline{=====}
\hline
\multicolumn{1}{c|}{\multirow{3}{*}{Model}}           & \multicolumn{2}{c|}{Trained Objects}                                                                                                          & \multicolumn{2}{c}{Untrained Objects}                                                                                                         \\ \cline{2-5} 
\multicolumn{1}{c|}{}                                 & Complete Success Rate {[}\%{]},                                        & Partial Success Rate {[}\%{]},                                         & Complete Success Rate {[}\%{]},                                        & Partial Success Rate {[}\%{]},                                         \\
\multicolumn{1}{c|}{}                                 & Successful Trials/Total Trials                                        & Successful Trials/Total Trials                                        & Successful Trials/Total Trials                                        & Successful Trials/Total Trials                                        \\ \hline
\multirow{2}{*}{\begin{tabular}[c]{@{}l@{}}I : AE-LSTM + Constraint \\     + Attention\end{tabular}} & \multirow{2}{*}{\begin{tabular}[c]{@{}c@{}}84.4, 27/32\end{tabular}} & \multirow{2}{*}{\begin{tabular}[c]{@{}c@{}}100, 32/32\end{tabular}}  & \multirow{2}{*}{\begin{tabular}[c]{@{}c@{}}66.7, 24/32\end{tabular}} & \multirow{2}{*}{\begin{tabular}[c]{@{}c@{}}94.4, 34/36\end{tabular}} \\
                                                      &                                                                       &                                                                       &                                                                       &                                                                       \\
\multirow{1}{*}{II : AE-LSTM + Constraint}            & \multirow{1}{*}{\begin{tabular}[c]{@{}c@{}}78.1, 25/32\end{tabular}} & \multirow{1}{*}{\begin{tabular}[c]{@{}c@{}}87.5, 28/32\end{tabular}} & \multirow{1}{*}{\begin{tabular}[c]{@{}c@{}}63.9, 23/36\end{tabular}} & \multirow{1}{*}{\begin{tabular}[c]{@{}c@{}}72.2, 26/36\end{tabular}} \\
\multirow{1}{*}{III : AE-LSTM + Attention}            & \multirow{1}{*}{\begin{tabular}[c]{@{}c@{}}62.5, 20/32\end{tabular}} & \multirow{1}{*}{\begin{tabular}[c]{@{}c@{}}96.9, 31/32\end{tabular}} & \multirow{1}{*}{\begin{tabular}[c]{@{}c@{}}36.1, 13/36\end{tabular}} & \multirow{1}{*}{\begin{tabular}[c]{@{}c@{}}80.6, 29/36\end{tabular}} \\
\multirow{1}{*}{IV : AE-LSTM}                         & \multirow{1}{*}{\begin{tabular}[c]{@{}c@{}}62.5, 20/32\end{tabular}} & \multirow{1}{*}{\begin{tabular}[c]{@{}c@{}}81.3, 26/32\end{tabular}} & \multirow{1}{*}{\begin{tabular}[c]{@{}c@{}}36.1, 13/36\end{tabular}} & \multirow{1}{*}{\begin{tabular}[c]{@{}c@{}}69.4, 25/36\end{tabular}}                                                                                      \\ \hline
\hhline{=====}
\end{tabular}
\end{table*}

   \begin{figure}[t]
      \centering
      \includegraphics[scale=0.4]{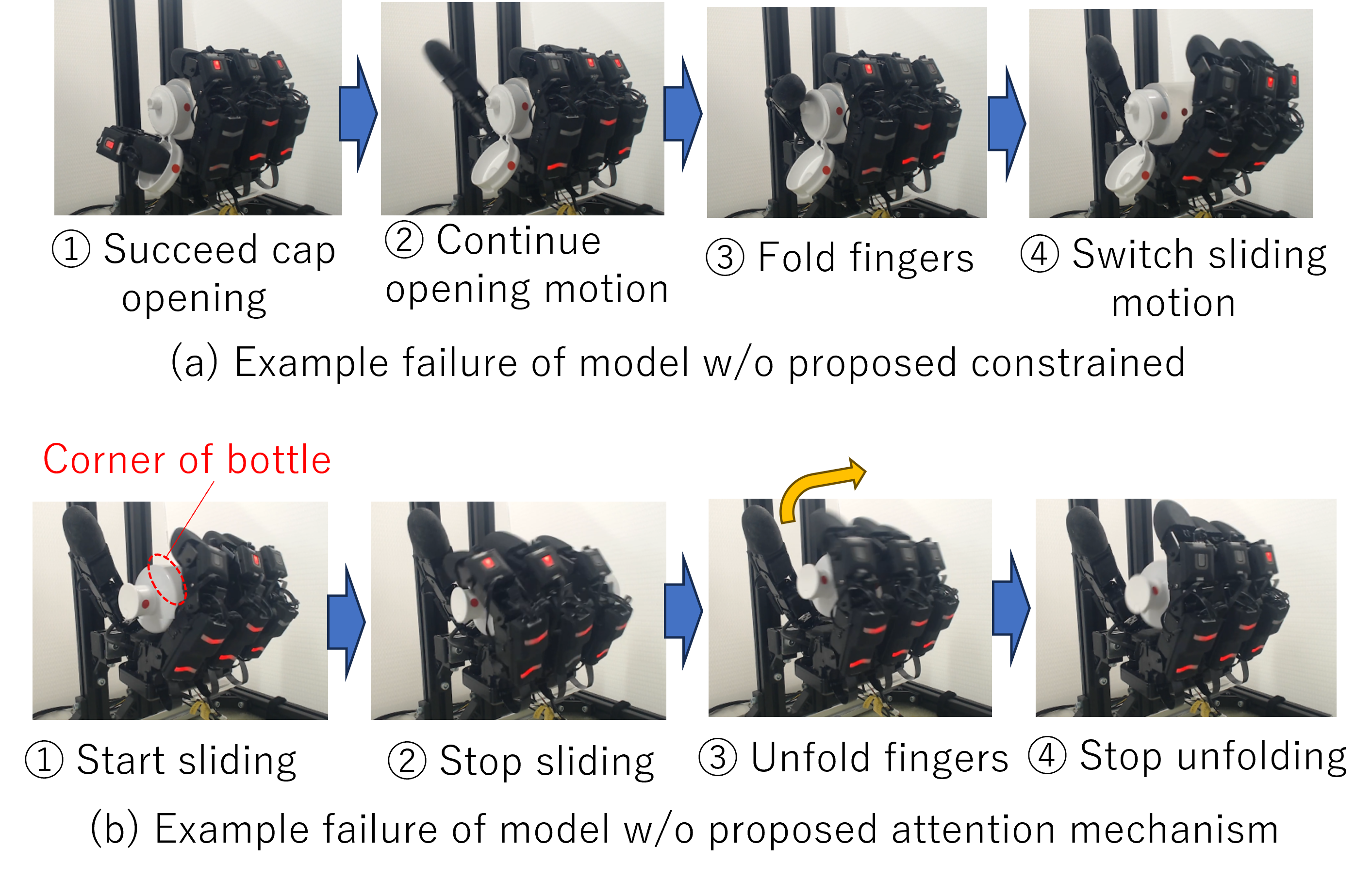}
      \caption{Failures made by comparison models. Top row shows when the model without constrained generates motion. It could not recognize that the cap was opened. Bottom row shows when the model without attention mechanism generates motion. It could not slide the object enough and thumb got stuck at the rim of the object and the opening motion failed.}
      \label{faliures}
   \end{figure}

\subsubsection{Evaluation Setting}
To evaluate the proposed constraints on the loss function and an attention mechanism,
we prepared three types of baseline model. The first model, AE-LSTM, uses only constraints on the loss function; the second model, AE-LSTM, utilizes only the attention mechanism; and the third model, AE-LSTM without employing the proposed method, 
for comparing the success rate of cap-opening motions. 
Moreover, we quantitatively evaluate the partial success rate and complete success rate of cap-opening motions. Here are the definitions:

\begin{itemize}
\item Complete success: If the cap is opened and recognized, and the motion is immediately stopped, it is considered a complete success.
\item Partial Success: If the cap-opening motion is successful but not recognized, resulting in the continued execution of the motion, it is considered a partial success.
\end{itemize}

The duration of the motion evaluation experiment was limited to 90 seconds. Failure was recorded if the cap-opening motion was not successful within this time frame (the maximum motion time during training data collection was approximately 57 seconds).

Firstly, to assess positional generalization, we evaluate the success rate of the target manipulation using four initial positions
with trained objects (yellow circles in Fig. \ref{datasetting}). Secondly, to evaluate object generalization, we assess the success rate of the target manipulation with untrained objects at the three testing initial positions shown in Fig. \ref{datasetting}. We conducted the manipulation two times for each combination of object and initial position. Therefore, for the trained objects, a total of 32 trials were conducted (4 trained objects x 4 testing initial positions x 2 trials), while for untrained objects, a total of 36 trials were conducted (6 untrained objects x 3 testing initial positions x 2 trials).

\subsubsection{Experiment Results}
The results of the evaluation experiments are presented in Table \ref{tab:3}. The proposed AE-LSTM model (model I) using constraints on the loss function and an attention mechanism achieved the highest partial success rate and complete success rate. This indicates that the proposed method has acquired generalization performance for initial positions and objects. By imposing constraints on the loss function (model II), it was possible to switch to a stop motion immediately after the cap was opened, leading to an improvement in the complete success rate. On the other hand, for models (III and IV) without constraints, even if the cap was opened, they often continued the motion without switching to a stop. Additionally for model I and III, by focusing attention on appropriate modalities for each motion 
the accuracy of motion was enhanced, resulting in improved partial success rates for both sliding and cap-opening motions. Conversely, for models (II and IV) without the attention mechanism, the accuracy of sliding motions was low, leading to frequent failures in cap-opening due to the inability to move the object smoothly left and right. 
From those results 
it can be inferred that constraints on the loss function contributed to improving the complete success rate, while the attention mechanism contributed to increasing the partial success rate. Therefore, the proposed method, which combines both constraint and attention techniques, could achieve the most successful in-hand manipulation.



\subsection{Analysis on Attention Value}
   \begin{figure}[b]
      \centering
      \includegraphics[scale=0.50]{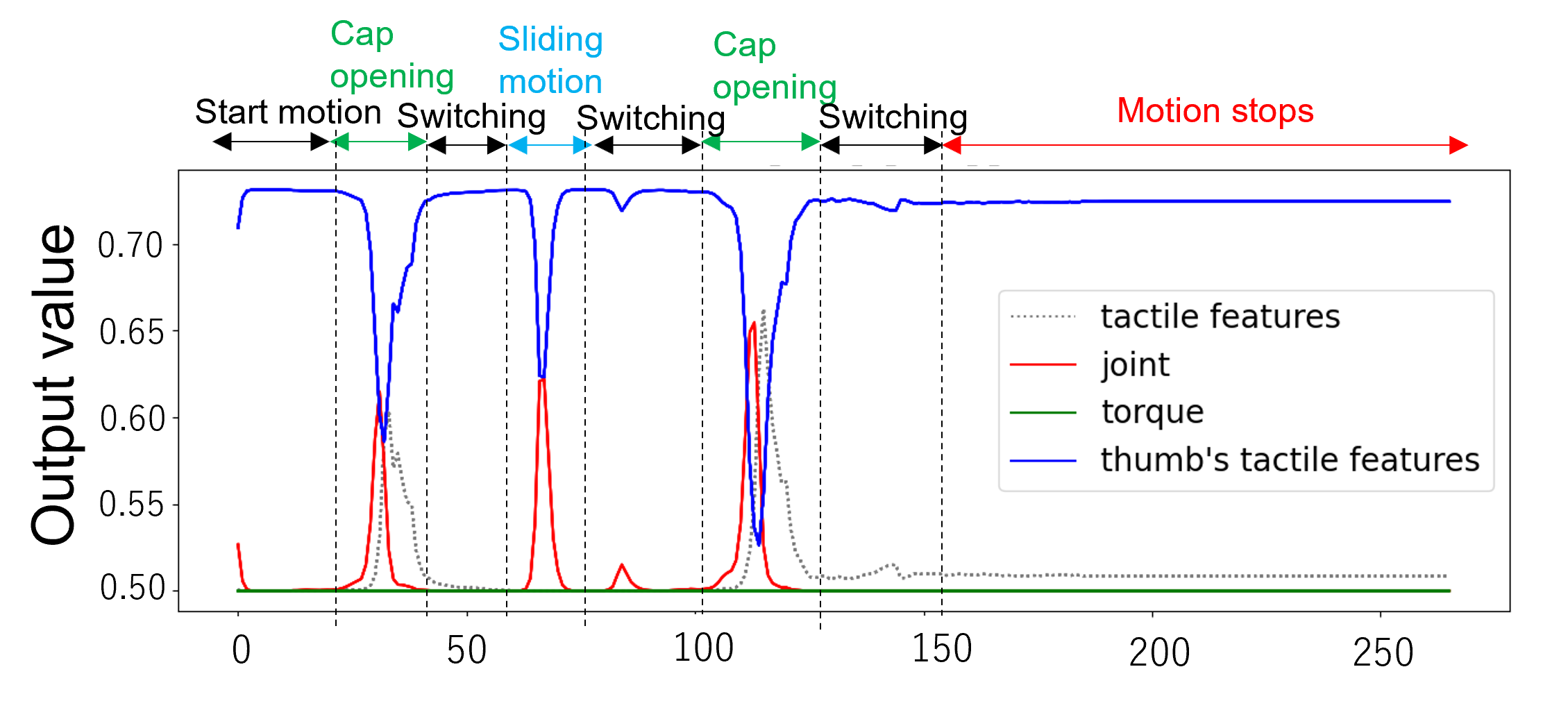}
      \caption{Trajectories of attention values for joint, tactile and torque values from the entire Allegro Hand for one single manipulation trial.
      }
      \label{attention}
   \end{figure}
To assess whether the attention mechanism appropriately switches attention to modalities during motion generation, we visualized the weighting values for each modality
by the attention mechanism in the proposed model as shown in Fig. \ref{attention}. The horizontal axis represents the time at intervals of 10Hz, and the vertical axis represents the output values of the attention mechanism. 
During the cap-opening motion, attention is directed to joint angles and whole tactile features. Throughout the sequence of motions, the object's posture may sometimes be parallel to the hand and other times diagonal, causing the position for opening the cap to be difficult. Therefore, it is presumed that attention to these modalities during the cap-opening motion is for accurately capturing the position to open the cap. 
Additionally, immediately after the cap-opening motion, attention is focused on the tactile features of the thumb. This seems to be for confirming the state of the cap in order to switch motions. Furthermore, during the sliding motion, attention is directed to joint angles. As shown in Fig. \ref{faliures}, if the accuracy of sliding motion is low, it is difficult to move the object and the motion tends to fail. This demonstrates that focusing attention on joint angles enables precise motion. 
These results imply that the attention mechanism appropriately directs attention to important modalities for each sub-task.

\subsection{Analysis on Features of LSTM}
   \begin{figure}[t]
      \centering
      \includegraphics[scale=0.82]{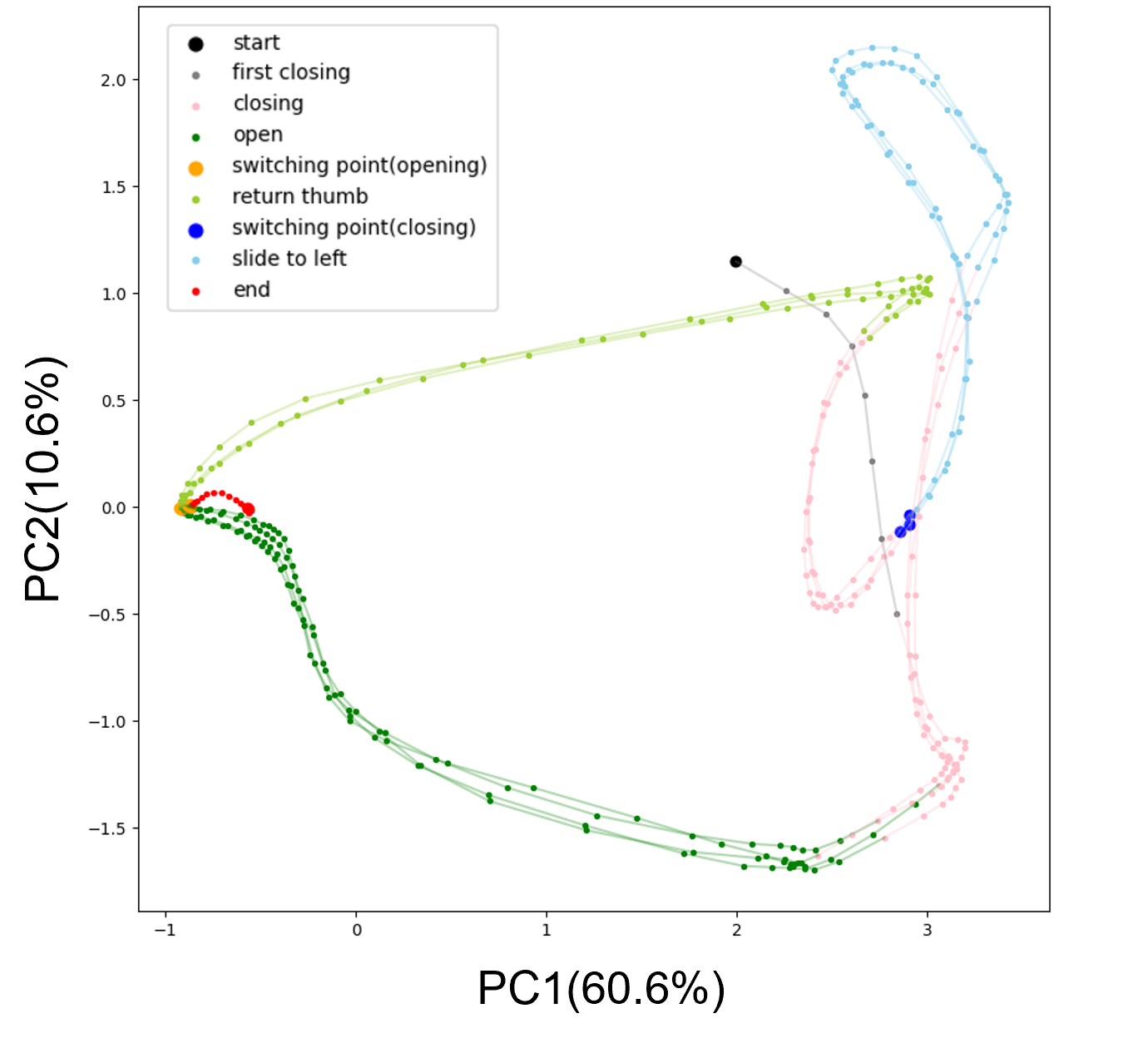}
      \caption{A PCA map of features from the LSTM. 
      Clusters for the phase of each motion clearly emerge. This map is for the case of sliding motion toward left. }
      \label{pca}
   \end{figure}
To analyze the effect of constraints on the proposed loss function, we investigated the latent variables of LSTM in the proposed model through principal component analysis (PCA) and analyzed whether the motions for sub-tasks were embedded in the latent space.
The horizontal and vertical axis represents the first and the second principal component, respectively.
Each plot connects the transition of latent variables at each timestep from the start to the end of the task with a straight line. 
As shown in Fig. \ref{pca}, imposing constraints on the loss function forced the latent variables of LSTM to trace a loop trajectory with the timing of motion switching. This suggests that the proposed constraints
works and has contributed to improving the ability to switch motions. 
With the cap-opening motion positioned to the left (dark and light green dots) and the sliding motion to the right (light blue dots) on the horizontal axis, it can be inferred that 
the cap-opening manipulation was achieved stably by the model 
which could embed these two main motions in different latent spaces clearly.

\section{CONCLUSIONS}
This study proposed
an AE-LSTM model that combined restrictions on loss functions to improve the ability to switch multi-fingered motions and an attention mechanism that modulated attention to modalities including tactile and joint information for each sub-task.
We confirmed that the proposed method exhibited the highest success rate of in-hand manipulation with unlearned initial positions and with unlearned objects. As a future work, auto-encoders which consider spatial processing of tactile information such as \cite{funabashigcn} can be useful 
to be more adaptive to fine manipulation. 

\bibliographystyle{IEEEtran}
\bibliography{reference}

\end{document}